\documentclass[sigconf]{acmart}
\AtBeginDocument{%
  \providecommand\BibTeX{{%
    Bib\TeX}}}

\setcopyright{acmlicensed}
\copyrightyear{2018}
\acmYear{2018}
\acmDOI{XXXXXXX.XXXXXXX}
\acmConference[Conference acronym 'XX]{Make sure to enter the correct
  conference title from your rights confirmation email}{June 03--05,
  2018}{Woodstock, NY}
\acmISBN{978-1-4503-XXXX-X/2018/06}

\usepackage{enumitem}
\usepackage{tabularray}
\usepackage{lipsum} %
\usepackage{graphicx} %
\usepackage{multirow}
\usepackage{CJKutf8}
\usepackage{hyperref} %

\begin{document}

\title{AIstorian lets AI be a historian: A KG-powered multi-agent system for accurate biography generation}

\author{Fengyu Li$^{\dagger}$, Yilin Li$^{\dagger}$, Junhao Zhu$^{\dagger}$, Lu Chen$^{\dagger}$, Yanfei Zhang$^{\dagger}$, Jia Zhou$^{\dagger}$, \\ Hui Zu$^{\dagger}$, Jingwen Zhao$^{\ddagger}$, Yunjun Gao$^{\dagger}$}
\affiliation{%
  \institution{$^{\dagger}$Zhejiang University, $^{\ddagger}$Poisson Lab, Huawei\\ $\{$fengyuli, zhujunhao, luchen, yanfei.zhang, 0012802, zdzh, gaoyj$\}$@zju.edu.cn, yilin.23@intl.zju.edu.cn, jingwenzhao5@huawei.com}
  \country{}
}

\renewcommand{\shortauthors}{Li et al.}

\begin{abstract}
Huawei has always been committed to exploring the AI application in historical research.  Biography generation, as a specialized form of abstractive summarization, plays a crucial role in historical research but faces unique challenges that existing large language models (LLMs) struggle to address. These challenges include maintaining stylistic adherence to historical writing conventions, ensuring factual fidelity, and handling fragmented information across multiple documents. We present \textsf{AIstorian}, a novel end-to-end agentic system featured with a knowledge graph (KG)-powered retrieval-augmented generation (RAG) and anti-hallucination multi-agents. Specifically, \textsf{AIstorian} introduces an in-context learning based chunking strategy and a KG-based index for accurate and efficient reference retrieval. Meanwhile, \textsf{AIstorian} orchestrates multi-agents to conduct on-the-fly hallucination detection and error-type-aware correction. Additionally, to teach LLMs a certain language style, we finetune LLMs based on a two-step training approach combining data augmentation-enhanced supervised fine-tuning with stylistic preference optimization. Extensive experiments on a real-life historical Jinshi dataset demonstrate that \textsf{AIstorian} achieves a $3.8\times$ improvement in factual accuracy and a $47.6\%$ reduction in hallucination rate compared to existing baselines. The data and code are available at: \href{https://github.com/ZJU-DAILY/AIstorian}{https://github.com/ZJU-DAILY/AIstorian}.
\end{abstract}

\maketitle

\section{Introduction}
Huawei has a commercial partnership with several museums, exploring the application of AI in the management of historical documents. Biography generation is an important intersection of AI and history. Typically, biography writing relies heavily on the domain expertise of historians who carefully gather and summarize diverse historical materials. The development of automatic biography generation can free historians from tedious tasks of reading and writing materials.
Biography generation automatically writes a concise document comprising a series of facts and events (e.g., date and place of birth and death, career, etc.) that builds a professional image for a person's life, by extracting and rephrasing biographical events based on references (e.g., textbooks). 
Biography generation can be considered a special form of abstractive summarization~\cite{Abstractive-Text-Summarization}.
Recent studies~\cite{pu2023summarizationalmostdead, Abstractive-Text-Summarization} have demonstrated the superior performance of LLMs for abstractive summarization, manifesting significant potential of LLMs in biography generation.
However, distinct from general text summarization, the biography generation exhibits unique characteristics that neither existing LLM-based methods nor vertical pretrained models can accommodate:

\noindent\textbf{Characteristic 1: Stylistic adherence.} Biographies have to strictly follow specific language styles and use professional terminology, which LLMs can struggle with. For example, LLMs cannot keep consistency when required to write a biography in classical Chinese, especially following historical writing conventions. To address this, several work injected domain expertise into LLMs via fine-tuning~\cite{DBLP:longt5,DBLP:pegasus}. Due to the scarcity of training corpora that follow a particular style, however, fine-tuned LLMs still struggle to adhere to specific language styles and professional terminology.

\begin{figure*}[t]  %
    \centering
    \vspace{-3mm}
    \includegraphics[width=\textwidth, height=2.7in]{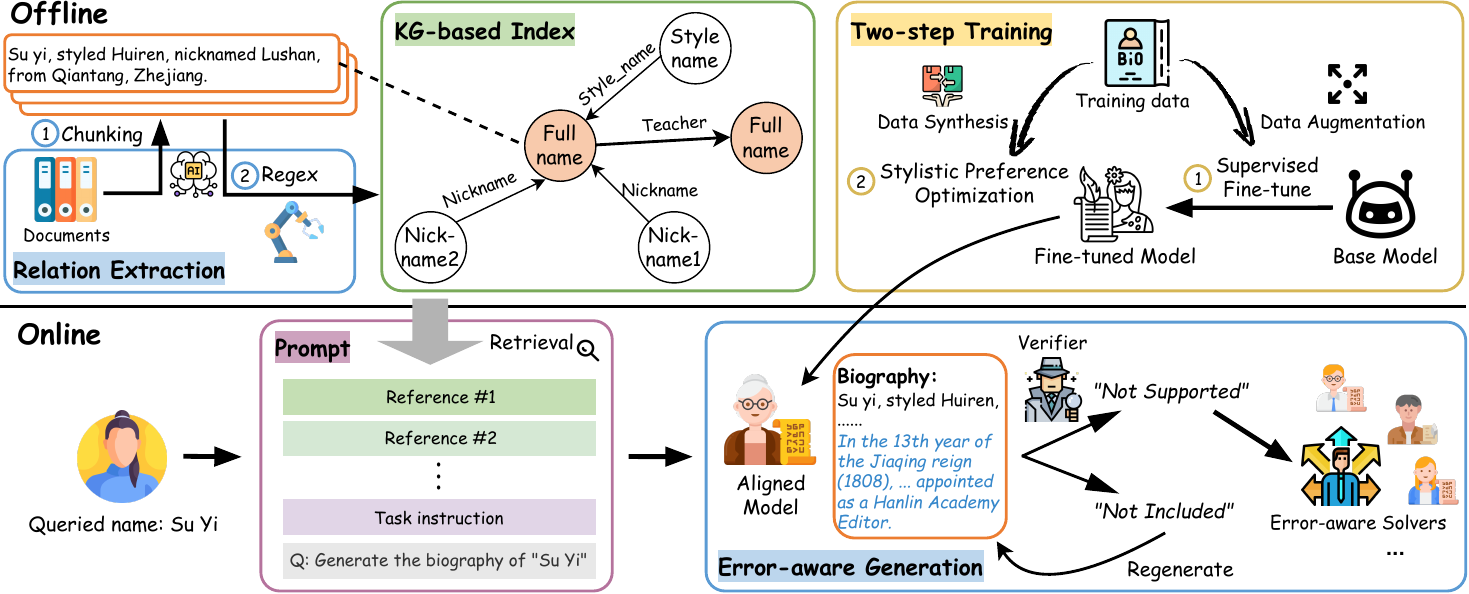}  %
    \caption{The architecture of \textsf{AIstorian}}
    \vspace{-3mm}
    \label{fig:architecture}
\end{figure*}

\noindent\textbf{Characteristic 2: Factual fidelity.} Biographies have no tolerance for factual errors. However, since LLMs are notorious for the phenomenon of hallucination, they would generate plausible content that contradicts real facts, significantly undermining the usability of generated biographies~\cite{min-etal-2023-factscore, FactualConsistencyNewsSummarization, Abstractive-Text-Summarization,liu-etal-2024-benchmarking}. There are three kinds of work proposed to alleviate fabrications in text summarization: (1) Training-based methods~\cite{DBLP:longt5,LLMsKnowMoreThanTheyShowIntrinsicRepresentation,chuang-etal-2024-lookback} fix factual errors inherently by training additional reward models or classifiers. A key drawback is their reliance on tens of thousands of training samples. (2) Decoding-based methods~\cite{menick2022teachinglanguagemodelssupportverifiedquotes} avoid hallucinations by constraining output tokens sampled from the given contexts. Though effective, such methods do harm to the model's creativity. (3) Retrieval-augmented-generation (RAG) based methods~\cite{yu-etal-2024-truthTruth-AwareContextSelection,edge2024localgraphrag} guide biography generation for LLMs by retrieving relevant information from an external knowledge base (e.g., history textbooks) as references. Apparently, the retrieval accuracy makes a pivotal difference in the quality of LLMs' generation.

\noindent\textbf{Characteristic 3: Information fragmentation.} Detailed information about a particular historical figure is scattered across multiple documents. Text chunking, which splits long documents into small text chunks, and chunk representation, which embeds text chunks into vectors, are two key components in a RAG workflow, suffering from the information fragmentation due to two-fold reasons: (1) The varied length of text to describe a single person poses challenge for text chunking, introducing irrelevant noisy information into a text chunk. (2) Aliases and pronouns in sentences make semantics vague, so that semantic embeddings of text chunks become indistinguishable, leading to the problem of representation dilution~\cite{liu2024retrieval-attention-acceleratinglongcontextllm}.

Towards accurate biography generation, this work presents \textsf{AIstorian}, a novel agentic system that consists of a knowledge graph (KG)-powered RAG mechanism and anti-hallucination multi-agents. Specifically, in the KG-powered RAG, we design an in-context learning based chunking strategy and build a KG-based index for the external knowledge base, which facilitates efficient and accurate retrieval through information reorganization. To further alleviate LLM's hallucinations, our system incorporates real-time anti-hallucination multi-agents, which involve on-the-fly hallucination detection and error-type-aware hallucination correction, ensuring that the generated biographies remain faithful to the source documents. Moreover, to better accommodate to specific language styles, we fine-tune the base model via a two-step training process including data augmentation-enhanced supervised fine-tuning and stylistic preference optimization.

The key contributions of this work are summarized as below:
\begin{itemize}[topsep=0pt, leftmargin=*]
    \item \emph{A novel end-to-end agentic system for biography generation}. Towards accurate biography generation, we design a novel end-to-end agentic system comprising a KG-powered RAG mechanism and anti-hallucination multi-agents. To the best of our knowledge, this is the first work to generate historical biographies from a massive corpus.
    \item \emph{An efficient KG-powered RAG mechanism}. To search for references efficiently and accurately, we present a KG-powered RAG mechanism, which reorganizes the knowledge base via an in-context learning based text chunking and an an efficient KG-based index.
    \item \emph{Error-aware anti-hallucination multi-agents}. To keep high fidelity, we design error-aware anti-hallucination multi-agents that integrate a fine-tuned LLM with real-time hallucination detection and error-aware correction mechanisms.
    \item Through extensive experiments on the Jinshi dataset, we demonstrate that our system achieves superior performance in both generation quality (Rouge-L: 80.54) and factual accuracy, with a 47.6\% reduction in hallucination rate and a 3.8$\times$ decrease in average atomic fact error.
\end{itemize}
\section{\textsf{AIstorian}}

\subsection{Overview}

Figure~\ref{fig:architecture} illustrates the overall architecture. The system operates in two modes, with upper half of the figure showing the offline process of index building and model fine-tuning, and the lower half showing the online biography generation pipeline.
During the offline phase, \textsf{AIstorian} completes two preparations: (1) KG-based index construction, which reorganizes the biographical facts of historical figures into a knowledge graph, and (2) two-step training, which fine-tunes and aligns the base model on small handcrafted biographical data to teach LLMs a specific language style.
During the online phase, \textsf{AIstorian} takes as an input a query of ``generating a biography of a historical figure'', and then invokes the KG-powered RAG to retrieve relevant information (e.g., text chunks) with the assistance of KG-based index. Using the query and the retrieved relevant information, the aligned LLM starts biography generation. Meanwhile, a Verifier validates the correctness of the generated content in real time. Once an error is detected, a Router categorizes the error and invokes the Solvers to correct it.

\subsection{KG-based RAG}

\subsubsection{Offline KG-based index construction} 
\begin{figure}[t]  %
    \centering
    \includegraphics[width=1\columnwidth]{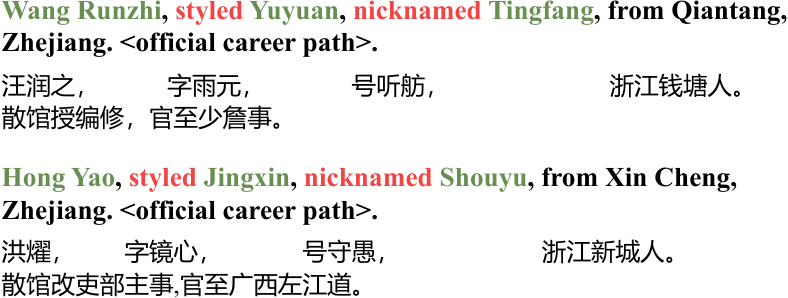}  %
    \caption{Examples of writing patterns (from the book \textit{Forest of Words: Compilation} written in classical Chinese).}
    \label{fig:pattern}
    \vspace{-3mm}
\end{figure}

Documents within the same book series reveal consistent structural patterns: the introduction of person entities follows a standardized format. As illustrated in Figure~\ref{fig:pattern}, the introduction of historical figures always begins with the name, followed by the alias and other details (e.g., hometown, career, etc.). Motivated by these, we propose a three-step index construction process to effectively reorganize biographical facts.

\noindent\textbf{Training-free pattern-enhanced text chunking}. Considering that most documents in the external knowledge base follow the same writing pattern, we can leverage the pattern to split long documents into smaller text chunks using in-context learning~\cite{DBLP:Bayesian-Inference-ICL}. Specifically, we use the pattern as instructions to guide the LLM in splitting the documents. Moreover, to enhance accuracy, several chunks following the pattern are handcrafted as demonstrations for the LLM. In this way, we achieve flexible and accurate text chunking based on the power of in-context learning without requiring expensive model training~\cite{chen-etal-2024-Dense-retrieval}.

\noindent\textbf{Regex-driven relation extraction}.
In this step, we aim to extract biographical facts from text chunks in the form of triplets (comprising a head entity, a relation, and a tail entity). To balance accuracy and flexibility, rather than extracting triplets directly using the LLM, we prompt the LLM to generate a specific regex for each chunk and apply the regex to the text chunk. Taking Figure~\ref{fig:pattern} as an example, with the LLM, we have a regex like ``\textup{\textsf{\small (\textbackslash{}S+), styled (\textbackslash{}S+), nicknamed (\textbackslash{}S+)}}''. To provide additional knowledge for the LLM, we use handcrafted regex and high-precision LLM-generated regex as demonstrations. In case where the LLM fails to generate an effective regex, we resort to plan B, which involves extracting triplets directly using the LLM.

\noindent\textbf{Knowledge graph construction}.
Finally, the biographical knowledge graph is built by treating entity names as nodes and relationships between entities as edges. Each node in the KG is linked to its source text chunk where the triplets were extracted, and one can navigate to the text chunk by searching for nodes in the KG.

\subsubsection{Online retrieval}
Upon receiving the query ``generate a biography of someone'',  \textsf{AIstorian} identifies the node of interest by traversing the knowledge graph, and retrieves the text chunks associated with the nodes as references. To expand knowledge coverage, we also consider neighboring nodes as nodes of interest, which are often overlooked by prior methods~\cite{cao2024tonggu, Atlas}.

\subsection{Error-aware Generation with Multi-agents}

\subsubsection{Offline LLM re-training} Directly applying general-purpose LLMs falls short in terms of biography generation quality. Meanwhile, the scarcity of professional biographical training data limits the capability of fine-tuning LLMs~\cite{AttrPrompt, lee-etal-2024-llm2llm}. To address this, we propose a two-step training strategy, including supervised fine-tuning (SFT) and stylistic preference optimization (StylePO), to enhance the LLM's performance with a limited amount of training data.

\noindent\textbf{Data augmentation-enhanced supervised fine-tuning.} To overcome data scarcity, we propose a biography data augmentation strategy, involving two phases: (1) Shuffle the key biographical facts in the given references to improve the LLM's ability to extract key information, and (2) introduce ``distractor documents''~\cite{zhang2024raft} (i.e., documents containing irrelevant noise) into the references to improve the LLM's ability to extract correct information.

\noindent\textbf{Stylistic preference optimization (StylePO).} To further align the LLM with preferences for high factual fidelity and a classical Chinese language style, we adopt the SimPO algorithm~\cite{meng2025simpo} to proceed fine-tuning the base model. The training set consists of pairs of positive and negative samples. In our setting, each positive sample is a golden biography written by experts in classical Chinese, while each negative sample is generated by translating the positive sample into modern Chinese and adding (or removing) additional details to (or from) the biography. SimPO steers the LLM to more likely generate biographies whose style resembles the expert-written ground truth.

\subsubsection{Online biography generation.} Offline model re-training improves generation quality, but it is inevitable that the LLM will output plausible content due to its inherent hallucination nature. Moreover, since LLMs output in an auto-regressive manner, errors in LLM outputs have a cascading effect, where one error results in further errors in LLM generation~\cite{chuang-etal-2024-lookback}. Therefore, error correction should be executed alongside biography generation. To this end, \textsf{AIstorian} orchestrates seven agents to automatically discover and fix errors during LLM generation. The multi-agents consist of a Verifier, a Router, and five Solvers.

\noindent\textbf{Verifier.} Following prior work~\cite{min-etal-2023-factscore}, we define the ``atomic fact'' as a short sentence conveying one piece of biographical information. The core logic of the Verifier is to check whether each atomic fact in the generation can be supported by the retrieved references. To achieve this, the Verifier first filters out the irrelevant documents from all retrieved references to reduce the number of references to be considered, based on the token-level Jaccard similarity between the atomic fact and each reference. In the following, the Verifier instructs the LLM to extract information related to the atomic fact from the remaining references and checks whether the atomic fact is supported. To avoid the cascading effect, the Verifier is executed for every single sentence generated. Once an error is found, \textsf{AIstorian} invokes the subsequent Router and Solvers.

\noindent\textbf{Router.} Based on observations, the errors occurred in the generations are classified into two main categories:
\begin{itemize}[topsep=0pt, leftmargin=*]
    \item \emph{``Not-included'' error}. A ``not-included'' error occurs when the erroneous atomic fact is not founded in the given references.
    \item \emph{``Not-supported'' error}. A ``not-supported'' error occurs when the atomic fact in the generation contradicts the given references. ``Not-supported'' errors can be classified into finer-grained error types based on the content of the atomic fact.
\end{itemize}

The Router determines the solutions to correct the detected errors based on their types. For ``not-included'' errors, \textsf{AIstorian} fixes these errors by regeneration; for ``not-supported'' errors, \textsf{AIstorian} invokes Solvers to provide more reliable solutions.

\begin{table*}[tb]\small
\centering
\caption{The biography generation performance}
\vspace{-4mm}
\renewcommand{\arraystretch}{1.}
\setlength{\tabcolsep}{2mm}{
\begin{tabular}{ l  @{}l  @{} c  c  c  c  c}
\toprule[1pt]
& \textbf{Model}                 & \textbf{ROUGE-1↑} & \textbf{ROUGE-2↑} & \textbf{ROUGE-L↑} & \textbf{Hallc-Rate↓} & \textbf{Average Atomic Fact Error↓} \\ \hline
 \multirow{2}{0.13\textwidth}{\textbf{Long Document Summary}} &  
mLongT5\_large & 41.38  & 17.69  & 28.08  & 96.43    & 5.29                                \\
& Qwen2.5-72B-Instruct-AWQ & 39.43    & 21.37   & 33.76  & 96.43    & 3.41                                \\ \hline
\multirow{4}{*}{\textbf{RAG+Summary} } &  
TongGu     & 69.64         & 56.48      & 65.90      & 89.29         & 1.96                                \\
& GraphRAG   & 36.45      & 18.69      & 30.53      & 96.43     & 5.21                           \\
& CL-KL (NER) & 73.76       & 59.16      & 68.46    &  92.86        & 1.96                                    \\
&  BI-LSTM-CRF (NER)   & 75.12   & 60.96   & 70.45   &  75.0           & 1.64                                    \\ \hline
\multirow{3}{*}{\textbf{Ours} } 
&   \textsf{AIstorian} & \textbf{83.69}  &\underline{74.14}  &\underline{80.54}  & \textbf{39.29}    & \textbf{0.43}
\\
 & w/o. Multi-agents & \underline{83.03}    & \textbf{74.34} & \textbf{81.16}  &\underline{57.14}  & \underline{1.14}    
\\
 &  w/o. StylePO \& Multi-agents     & 81.93  & 71.71 & 78.88 & \underline{57.14} & 1.75 
\\
\bottomrule[1pt]
\end{tabular}}
\label{tab:summary}
\vspace{-4mm}
\end{table*}

\noindent\textbf{Solver.} To fix various types of errors,  a bunch of Solvers is used. %
\begin{enumerate}[topsep=0pt, leftmargin=*]
    \item \emph{Era-conflict Solver}. For errors related to the timing of events, the Solver invokes the Python function to convert the era name into the Gregorian calendar and performs a double-check.
    \item \emph{Ref-conflict Solver}. For conflicts that occur in the references, the Solver involves expert intervention to resolve the conflict. 
    \item \emph{Knowledge-lack Solver}. For errors caused by a lack of expertise, the Solver retrieves information (e.g., imperial competitive examination, ancient geography, etc.) online to prompt the LLM.
    \item \emph{Alias-conflict Solver}. As a historical figure may have multiple aliases, the solver selectively presents all right aliases in the generated biography.
    \item \emph{Solver for others}. For other errors, the Solver fixes the error by prompting the LLM with error messages. 
\end{enumerate}

\section{Experiments}

\subsection{Experimental Settings}
\noindent\textbf{Dataset.}
We use the Jinshi dataset, a real-world historical dataset, for evaluation. It contains 173 Chinese historical figures with manually written biographies as ground truth and a knowledge base of ancient Chinese texts with approximately 220k Chinese characters.

\noindent\textbf{Baselines.} We consider the following six baselines:
\begin{enumerate}[topsep=0pt, leftmargin=*]
    \item \textbf{mLongT5\_large}~\cite{uthus-etal-2023-mlongt5}: A multilingual LongT5\_large~\cite{DBLP:longt5} designed for text summarization.
    \item \textbf{Qwen2.5-72B-Instruct-AWQ}~\cite{qwen2025qwen25technicalreport}: The SOTA open-sourced LLM with 72.7B parameters, featuring AWQ 4-bit quantization.
    \item \textbf{TongGu}~\cite{cao2024tonggu}: A RAG system combining a classical Chinese-specific LLM, a SOTA retrieval model (Conan-embedding-v1), and a reranker (bge-large-zh-v1.5).
    \item \textbf{GraphRAG}~\cite{edge2024localgraphrag}: A SOTA KG-enhanced RAG system for query-focused text summarization.
    \item \textbf{CL-KL(NER)}~\cite{wang2021improving}: A RAG system on top of Qwen2.5-7B-Instruct, where retrieval index is built using an named entity recognition (NER) model trained by cooperative learning.
    \item \textbf{BI-LSTM-CRF(NER)}~\cite{huang2015bidirectional}: A RAG system on top of Qwen2.5-7B-Instruct, where retrieval index is built by an NER model based on BI-LSTM networks and a CRF layer.
\end{enumerate}
For \textsf{mLongT5\_large} and \textsf{Qwen2.5-72B-Instruct-AWQ}, we retrieve relevant documents using our KG-based index due to the context length limitations.

\begin{table}[tb]\small
\centering
\caption{Retrieval performance}
\label{tab:retrieval}
\vspace{-3mm}
\begin{tabular}{c c c c}
\toprule[1pt]
\textbf{Model}             & \textbf{Precision} & \textbf{Recall} & \textbf{F1 Score}    \\  \hline
TongGu                         & 0.341             & 0.78            & 0.456                           \\
GraphRAG                       & 0.36         & 0.676        & 0.379                                \\
CL-KL (NER)         & 0.623          & 0.478             & 0.518                                \\
BI-LSTM-CRF (NER)   & 0.584            & 0.303	           & 0.359                          \\
Ours                 & \textbf{0.936} &\textbf{0.944} &  \textbf{0.923}                                \\
\bottomrule[1pt]
\end{tabular}
\vspace{-4mm}
\label{tab:retrieval}
\end{table}

\noindent\textbf{Implementation details.} 
Baseline parameters follow the original papers. The model training is conducted on an A40 46G GPU, while other experiments are run on an NVIDIA 4090 24G GPU. The RAG component uses LangChain~\cite{Chase_LangChain_2022} and the multi-agent system uses Qwen-Agent~\cite{qwen-agent}. We employ Qwen2.5-7B-Instruct~\cite{qwen2025qwen25technicalreport} as the foundation for both training model and multi-agents. In the SFT phase, the rank of LoRA is set to 32, the learning rate is 2e-4, the batch size is 8 and the model is trained for 2 epochs. In the StylePO phase, the rank of LoRA is set to 8, the learning rate is 5e-6, the batch size is 8, and the model is trained for 3 epochs. The dataset is split into a training set and a test set in the ratio of 8:2.

\noindent\textbf{Evaluation metrics.}
For retrieval, precision, recall and F1-score are used for evaluation. %
For biography generation, ROUGE-$n (n=1,2)$ evaluates generation quality by measuring the overlap of $n$-grams between generated text and ground truth, while ROUGE-L~\cite{lin-2004-ROUGE} evaluates quality by a longest common subsequence (LCS)-based F-measure.
Hallucination rate measures the proportion of generated text containing hallucinations, which are identified by human experts.
Following prior work~\cite{min-etal-2023-factscore}, we quantify generation accuracy using the average number of erroneous atomic fact, where the erroneous atomic fact is manually-labeled.

\subsection{Performance Comparison}
We compare our \textsf{AIstorian} with all baselines, followed by analyzing each components in the \textsf{AIstorian}.

\noindent\textbf{Overall performance}. Table~\ref{tab:summary} presents
the comparison of \textsf{AIstorian} and all baselines. As observed, \textsf{AIstorian} consistently outperforms all baselines, achieving $11.4\%$-$20.6\%$ improvements in biography generation quality and a $47.6\%$ reduction in hallucination rate compared to the previous SOTA methods.

\noindent\textbf{KG-based RAG.} We compare the KG-based RAG component with the RAG-based baselines. Table~\ref{tab:retrieval} lists the experimental results. Our KG-based RAG achieves a 50\% improvement in precision and a 21\% improvement in recall over the previous SOTA, demonstrating its superior retrieval efficacy.

\noindent\textbf{Error-aware biography generation.} Thanks to error-aware generation with multi-agents, \textsf{AIstorian} achieves better generation quality and a lower error rate than baselines. In addition, we show that each component plays different roles according to ablation studies. For example, multi-agents improve hallucination rates but negatively impacts creativity (measured by ROUGE scores).

\section{Conclusion}
In this paper, we propose \textsf{AIstorian}, an innovative KG-powered multi-agent system towards accurate historical biography generation, which is featured with a KG-powered RAG mechanism and error-aware biography generation with multi-agents. To the best of our knowledge, this is the first work to generate historical biographies from a massive corpus. The experimental results demonstrate that our system significantly improves the quality of biography generation and alleviates hallucinations.
\section*{Presenter \& Company Biography}
Fengyu Li is currently an intern at Huawei and a graduate student in the Software Engineering at Zhejiang University. His research interests mainly focus on RAG (Retrieval-Augmented Generation) and natural language processing. 

Huawei is a global leader in information and communications technology (ICT) solutions. It offers comprehensive services and solutions to help enterprises and developers achieve digital transformation and innovation, with its strong technical capabilities in cloud computing, AI, and big data.

\balance
\bibliographystyle{ACM-Reference-Format}
\bibliography{Main.bbl}

\begin{table*}[t]
    \centering
    \caption{Qualitative analysis of Biographies. \textcolor{blue}{Blue} and
 \textcolor{red}{red} colors are used to indicate "Not included" errors and "Not supported" errors, respectively. \textcolor{cyan}{Cyan} is used to indicate the content that historians have overlooked.}
    \label{tab:analysis}
    \begin{tabular}{p{0.98\textwidth}}
        \toprule[1pt]
        \begin{CJK*}{UTF8}{gbsn}\textbf{Historian}: 
        胡鄂，字卓庵。明崇祯三年（1630）二月十三日生，浙江绍兴府余姚县人。治《尚书》。…… 殿试第二甲第七十九名，赐进士出身。顺治十八年（1661）任江苏松江府推官。\end{CJK*} \\
        Hu E, styled Zhuo’an. Born on the thirteenth day of the second month in the third year of Chongzhen (1630) of the Ming dynasty, a native of Yuyao County in Shaoxing Prefecture, Zhejiang Province. Studied the \textit{Book of Documents}…… He ranked seventy-ninth in the second class, in the palace examination, and was granted the title of Jinshi. In the eighteenth year of Shunzhi (1661), appointed as the judicial officer of Songjiang Prefecture in Jiangsu Province. \\
        \hline
        \begin{CJK*}{UTF8}{gbsn}\textbf{Qwen2.5-72B-Instruct}: 胡鄂，字卓庵，浙江绍兴府余姚县人，生于\textcolor{red}{明崇祯五年（1632）}二月十三日。\textcolor{blue}{鄂少时好学}，尤精《书经》。......殿试列二甲第七十九名，赐进士出身，\textcolor{cyan}{观政于通政司}。顺治十八年（1661），鄂授\textcolor{red}{江南松江府推官}，\textcolor{blue}{勤于政事，治狱明察，颇有政声}。鄂在松江任内，\textcolor{blue}{以清廉著称，深得民心。鄂著述不多，然其治学严谨}，于《书经》颇有研究，\textcolor{blue}{为时人所重}。\end{CJK*}\\
        Hu E, styled Zhuo’an, was a native of Yuyao County in Shaoxing Prefecture, Zhejiang Province. He was born on the thirteenth day of the second month in the \textcolor{red}{fifth year of Chongzhen in the Ming dynasty (1632)}. \textcolor{blue}{In his youth, E was fond of learning}, and he was especially proficient in \textit{the Book of Documents}......During the palace examination, he was placed seventy-ninth in the second class and was granted the title of jinshi. \textcolor{cyan}{He then began his political career by serving in the \textit{Tongzhengsi}}. In the eighteenth year of the Shunzhi era(1661), E was appointed as \textcolor{red}{the judicial officer of Songjiang Prefecture in Jiangnan}. \textcolor{blue}{He was diligent in administrative affairs, discerning in handling legal cases, and gained considerable renown for his governance}. During his tenure in Songjiang, \textcolor{blue}{E became renowned for his integrity and won the hearts of the people. Though not prolific in writing, his scholarship was meticulous}, and he conducted extensive research on the Book of Documents, \textcolor{blue}{earning him great respect from his contemporaries}. \\
        \hline
        \begin{CJK*}{UTF8}{gbsn}\textbf{AIstorian}: 胡鄂，字卓庵。明崇祯三年（1630）二月十三日生，浙江绍兴府余姚县人。治《尚书》。......殿试第二甲第七十九名，赐进士出身，\textcolor{cyan}{通政司观政}。顺治十八年（1661）授江苏松江府推官。\end{CJK*}\\
        Hu E, styled Zhuo’an. Born on the thirteenth day of the second month in the third year of Chongzhen (1630) of the Ming dynasty, a native of Yuyao County in Shaoxing Prefecture, Zhejiang Province. Studied the \textit{Book of Documents}. ......…… He ranked seventy-ninth in the second class, in the palace examination, and was granted the title of Jinshi, and \textcolor{cyan}{appointed as an observer in the \textit{Tongzhengsi}}. In the eighteenth year of Shunzhi (1661), appointed as the judicial officer of Songjiang Prefecture in Jiangsu Province. \\
        \bottomrule[1pt]
    \end{tabular}
\end{table*}

\appendix

\section{Appendix}
\subsection{Case study}

As shown in Table.~\ref{tab:analysis}, \textsf{AIstorian} demonstrates a higher degree of fidelity to the source material, accurately capturing key historical details with remarkable stylistic consistency to golden ones. Additionally, through StylePO, it successfully captures contextual details occasionally overlooked by experts (marked in \textcolor{cyan}{cyan}). In contrast, while the biography produced by Qwen2.5-72B-Instruct appears more fluent and readable, it frequently introduces inaccuracies, such as unsupported factual errors (marked in \textcolor{red}{red}) and fabricated content (marked in \textcolor{blue}{blue}), which compromise the reliability of its outputs.

\subsection{Efficiency Evaluation}

\begin{table}[tb]
\centering
\caption{Time consumption in each stage}
\label{tab:time}
\vspace{-3mm}
\small
\begin{tabular}{c c c}
\toprule[1pt]
\textbf{Offline} & \textbf{Offline Training(s)} & \textbf{Index Construction(s)} \\  \hline
TongGu             & 480           & 26.69  \\
GraphRAG           & -             & 5279.65 \\
CL-KL (NER)        & 532           & 885.68 \\
BI-LSTM-CRF (NER)  & 532           & 1037.31 \\
Ours               & 3168          & 422.65 \\  
\hline\hline
\textbf{Online} & \textbf{Online Retrieval(s)}  & \textbf{Generate(s)}  \\  \hline
TongGu             & 169.101       & 114             \\
GraphRAG           & 1.68         &  386.16         \\
CL-KL (NER)        & 0.00283        & 114              \\
BI-LSTM-CRF (NER)  & 0.000869     & 114   \\
Ours               & 0.00146   & 1236 \\  
\bottomrule[1pt]
\end{tabular}
\vspace{-2mm}
\end{table}

Table~\ref{tab:time} presents the time consumption for each stage of the compared methods. In offline training, our system takes 3168s, exceeding other methods due to the additional StylePO stage. Ablation studies (comparing "w/o. StylePO \& Multi-agents" versus "w/o. Multi-agents" in Table~\ref{tab:summary}) reveal that StylePO enhances ROUGE scores and decreases atomic fact errors by 35\%. This additional training time, as a one-time investment, is justified by the significant performance gains achieved. For index construction, which is critical as it scales with the corpus size, our approach completes in 422.65s, substantially outperforming GraphRAG (5279.65s). The NER baselines (CL-KL and BI-LSTM-CRF) demand more time for this stage (885.68s and 1037.31s respectively) as they necessitate model training (with labeling time not included in these measurements). While embedding-based RAG (TongGu) exhibits the lowest time consumption (26.69s), it delivers significantly inferior retrieval quality as demonstrated in Table~\ref{tab:retrieval}.

In online retrieval, our system achieves millisecond-level speed (0.00146s), significantly faster than TongGu (169.101s) and GraphRAG (1.68s). Our generation stage takes longer (1236s, ~36.35s per biography) due to our error-aware generation with multi-agents approach. As evidenced in Table~\ref{tab:summary}, this time investment significantly reduces hallucination rate and improves factual accuracy while maintaining high ROUGE scores.

\end{document}